\documentclass[journal,twoside,web]{ieeecolor}
\usepackage{generic}
\usepackage{cite}
\usepackage{caption}
\usepackage{xcolor}
\usepackage{amsmath,amssymb,amsfonts}
\usepackage{graphicx}
\usepackage{algorithm}
\usepackage{algpseudocode}
\usepackage{hyperref}

\usepackage{textcomp}
\usepackage{graphicx}
\usepackage{textcomp}
\usepackage{tabularx}
\usepackage{multirow}
\usepackage{booktabs}
\usepackage{adjustbox}
\usepackage{stfloats} % Include this in your preamble
\usepackage{float}
\usepackage{tikz}
\usepackage{subcaption}
\usepackage{soul}
\def\BibTeX{{\rm B\kern-.05em{\sc i\kern-.025em b}\kern-.08em
    T\kern-.1667em\lower.7ex\hbox{E}\kern-.125emX}}
\begin{document}
\title{MPCNN: A Novel Matrix Profile Approach for CNN-based Sleep Apnea Classification}
\author{Hieu X. Nguyen,  Duong V. Nguyen, Hieu H. Pham, and Cuong D. Do
\thanks{This work was supported by the VinUni-Illinois Smart Health Center in VinUniversity.}
\thanks{Hieu X. Nguyen,  Duong V. Nguyen, Hieu H. Pham, and Cuong D. Do are with the VinUni-Illinois Smart Health Center, VinUniversity, Hanoi, Vietnam.}
\thanks{E-mail: \{21hieu.nx,21duong.nv,hieu.ph,cuong.dd@vinuni.edu.vn\} }
\thanks{Hieu X. Nguyen,  Duong V. Nguyen, Hieu H. Pham, and Cuong D. Do are also with the College of Engineering and Computer Science, VinUniversity, Hanoi, Vietnam.}
\thanks{Corresponding author: \textcolor{blue}{cuong.dd@vinuni.edu.vn} (Cuong D. Do) }}

\maketitle

\captionsetup[table]{
    justification=centering,
    labelsep=newline,
    labelfont={bf,color=cyan}
}

\begin{abstract}
Sleep apnea (SA) is a significant respiratory condition that poses a major global health challenge. Previous studies have investigated several machine and deep learning models for electrocardiogram (ECG)-based SA diagnoses. Despite these advancements, conventional feature extractions derived from ECG signals, such as R-peaks and RR intervals, may fail to capture crucial information encompassed within the complete PQRST segments. In this study, we propose an innovative approach to address this diagnostic gap by delving deeper into the comprehensive segments of the ECG signal. The proposed methodology draws inspiration from Matrix Profile algorithms, which generate an Euclidean distance profile from fixed-length signal subsequences. From this, we derived the Min Distance Profile (MinDP), Max Distance Profile (MaxDP), and Mean Distance Profile (MeanDP) based on the minimum, maximum, and mean of the profile distances, respectively. To validate the effectiveness of our approach, we use the modified LeNet-5 architecture as the primary CNN model, along with two existing lightweight models, BAFNet and SE-MSCNN, for ECG classification tasks.  Our extensive experimental results on the PhysioNet Apnea-ECG dataset revealed that with the new feature extraction method, we achieved a per-segment accuracy up to 92.11 \% and a per-recording accuracy of 100\%. Moreover, it yielded the highest correlation compared to state-of-the-art methods, with a correlation coefficient of 0.989. By introducing a new feature extraction method based on distance relationships, we enhanced the performance of certain lightweight models, showing potential for home sleep apnea test (HSAT) and SA detection in IoT devices. The source code for this work is made publicly available in GitHub: https://github.com/vinuni-vishc/MPCNN-Sleep-Apnea.
\end{abstract}


\section{Introduction}
\label{sec:introduction}
Sleep apnea is a form of sleep-disordered breathing characterized by temporary interruptions in breathing or reduced breathing amplitude during sleep, often leading to arterial hypoxemia, hypercapnia, and sleep fragmentation \cite{Pathophysiology}. Additionally, SA has been recognized as an independent risk factor for cerebrovascular disease (CVD), a leading cause of adult mortality and disability globally \cite{rdong}. Statistically, this disorder is significantly prevalent, affecting an estimated 200 million people worldwide \cite{ZhangJ}. However, among those with SA, 93\% of middle-aged women and 82 \% of middle-aged men exhibiting symptoms from significant to severe levels have gone undetected \cite{YoungT}. Therefore, finding a method for early diagnosis of SA is crucial to prevent further complications and improve the overall quality of life for affected individuals.

Polysomnography (PSG) is widely recognized as the clinical gold standard for diagnosing SA, owing to its diagnostic precision \cite{SFQuan}. However, PSG has limitations due to its complexity, high costs, lack of sleep assessment labs, and the requirement for overnight testing \cite{OLeBon, PLloberes}. As a result, researchers also developed a system namely portable monitoring (PM) or home sleep apnea test (HSAT) to keep track of people's sleep convenience at home. The electrocardiogram (ECG) signal has emerged as one of the most effective signals in PM, which demonstrates a strong correlation with SA events because the drop in oxygen saturation can lead to irregular heart rate changes \cite{AManiaci}. Recently, deep learning is commonly applied to ECG analysis due to its increased precision and efficiency in classification problems \cite{kle,Ribeiro20,thao}. There has been a significant focus on ECG-based SA detection in research, such as \cite{wang2019, chen2021,li2018,Tu22,feng2021}, particularly emphasizing features like R-peaks and RR intervals for classification. While these approaches have shown considerable promise, there was one major limitation: using only R-peaks and RR intervals might not capture all the important details found in the complete ECG waveform including 5 waves: P, Q, R, S, T and its segment. Moreover, relying only on R-peaks and RR intervals might overlook important parts of the signal, potentially missing key information that can help with an accurate diagnosis.

\begin{figure*}[htbp]
    \centering
    \includegraphics[width=0.85\textwidth]{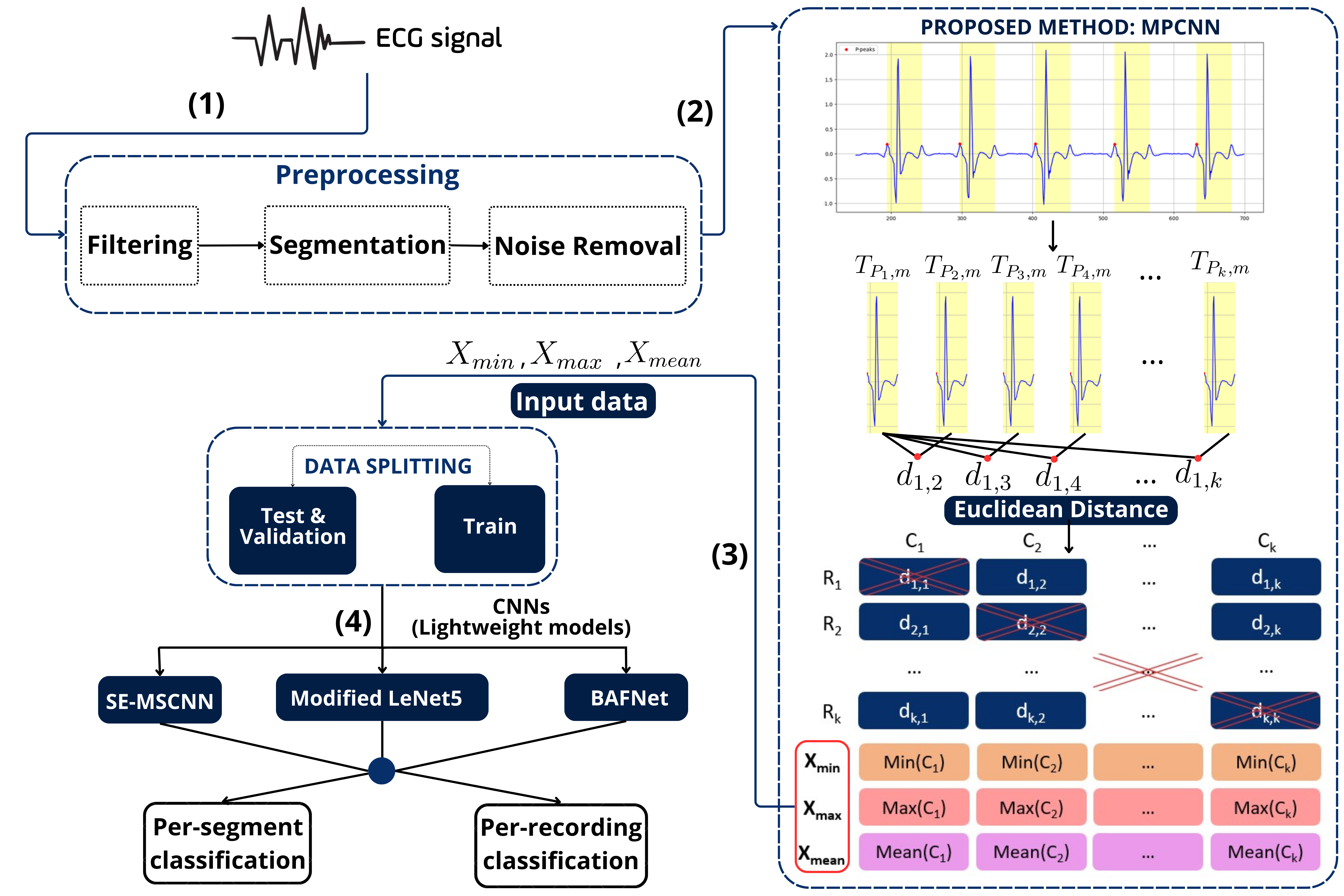}
    \caption{\textbf{The overall of the proposed approach SA detection}. Our approach contains 4 main components: (1) First, we reduce noise and artifacts from the original signals;  (2) Second, we generate a series of subsequences $T_{P_i,m}$, start at a P Peak and spanning a window of length $m$. We then calculate the Euclidean distance $d_{i,j}$ to compile a distance profile $D$, from which we extract critical values: $X_{min}$ (MinDP), $X_{max}$ (MaxDP), and $X_{mean}$ (MeanDP). These values serve as inputs for the subsequent modeling stage. (3) Next, we perform  data splitting in two categories; (4) Finally, we train some lightweight CNN-based models perform the per-segment ECG classification task.}
    \label{fig:my_label}
\end{figure*}
In this study, we introduce a novel feature extraction method named MPCNN for SA detection from a single-lead ECG signal. The Matrix Profile (MP) \cite{ChinChiaMichaelYeh} is a groundbreaking data structure for time series analysis, backed by powerful algorithms, that excels in delivering accurate, efficient, and scalable results. While primarily designed for the all-pairs-similarity-search challenge, incorporating its distance metrics into ECG analysis offers unprecedented depth in signal patterns. In particular, this method generates an Euclidean distance profile from fixed-length signal subsequences, effectively capturing the complete distance of the PQRST segments and its relationship of the whole signal through the value of MinDP, MaxDP and MeanDP, potentially offering a deeper insight into the ECG markers of sleep apnea. The visualization of this process is illustrated in phase signal processing of Fig. 1. To evaluate the proposed matrix representation, we conduct diverse experiments utilizing the standard CNN model modified LeNet-5 on PhysioNet dataset. We also evaluate the robustness of the performance by testing them with two distinct existing lightweight CNN models: BAFNet and SE-MSCNN. To sum up, this work's main contributions can be summarized as follows:
\begin{itemize}
    \item A novel feature extraction method based on MP algorithms is introduced, which goes beyond conventional ECG signal analysis by considering distance relationships within comprehensive signal segments, thereby improving the accuracy of SA detection.
    \item The proposed approach achieves per-segment accuracy of up to 92.11\% and per-recording accuracy of 100\%, outperforming state-of-the-art SA detection method. The method's robustness is demonstrated through testing with lightweight models: modified LeNet-5, BAFNet and SE-MSCNN.
    \item The results achieve the highest correlation coefficient, 0.989, when compared to state-of-the-art approaches, showcasing the effectiveness and reliability of the new feature extraction method.
    \item This study opens avenues for practical applications, including HSAT and SA detection in Internet of Things (IoT) devices, contributing to the advancement of accessible and effective healthcare solutions.
\end{itemize}

The remainder of this paper is structured as follows: Section II delves into related works. Section III details the proposed method, encompassing signal preprocessing, the new feature extraction based on the MP algorithms, a brief overview of some existing lightweight models, and performance evaluation metrics. Section IV outlines the experimental setup, presents the results with the discussion and the limitation of proposed methodology.

\section{Related Work}
% Use either SI (MKS) or CGS as primary units. (SI units are strongly 
% encouraged.) English units may be used as secondary units (in parentheses). 
% This applies to papers in data storage. For example, write ``15 
% Gb/cm$^{2}$ (100 Gb/in$^{2})$.'' An exception is when 
% English units are used as identifiers in trade, such as ``3\textonehalf-in 
% disk drive.'' Avoid combining SI and CGS units, such as current in amperes 
% and magnetic field in oersteds. This often leads to confusion because 
% equations do not balance dimensionally. If you must use mixed units, clearly 
% state the units for each quantity in an equation.

% The SI unit for magnetic field strength $H$ is A/m. However, if you wish to use 
% units of T, either refer to magnetic flux density $B$ or magnetic field 
% strength symbolized as $\mu _{0}H$. Use the center dot to separate 
% compound units, e.g., ``A$\cdot $m$^{2}$.''
\subsection{Home sleep apnea test for SA detection}

Instead of using PSG, employing HSAT offers an alternative approach for identifying SA, with the benefit of increased simplicity. Various physiological signs, including the electroencephalogram (EEG), electrocardiogram (ECG), and oxygen saturation (SpO2), have been identified as potential indicators for disease identification. In this study, ECG was employed as a diagnostic tool for identifying several cardiovascular conditions, including arrhythmia, atrial fibrillation, and others \cite{ygao, nihasan,fmurat}. Additionally, the detection of SA using ECG has been recognized as not only effective but also as a more accessible alternative compared to traditional methods \cite{mbahrami2021}.
\subsection{Feature extraction in SA classification}
Below, the feature extraction of the two groups of techniques most closely related to the proposed approach is discussed, including traditional machine learning-based approaches and deep learning-based approaches.
 
\textbf{Machine learning based methods:} Varon \textit{et al.} \cite{varon2015} developed a SA detection approach using novel ECG-derived features, including heart rate variability (HRV) and RRI (R-R interval) from extracted from ECG signal. Taking a different angle, Song \textit{et al.} \cite{song2015} introduced a distinctive hidden Markov model (HMM) for SA detection, specifically designed to address the issue of ignoring the temporal dependence commonly found in other models, providing a nuanced understanding of individual variations. Hassan \textit{et al.} \cite{hassan2016} introduced an automated approach for detecting SA using ECG signals that were decomposed using Empirical Mode Decomposition (EMD) in conjunction with intrinsic mode functions (IMF). Subsequently, a set of statistical features was chosen and incorporated into an extreme machine learning model. Besides that, 
Sharma \textit{et al.} \cite{sharma2016} pinpoints apnea segments in ECGs by leveraging Hermite coefficients combined with R-R time series features, and the efficacy of their method is tested across four distinct classifiers.

Most of the previously mentioned methods for SA detection involve two steps. The initial step involves gathering relevant data, while the second step entails selecting an appropriate classifier for the automated process. However, these methods suffer from certain drawbacks. In the first step, there exists an overwhelming number of features that can be chosen, leading to a potential issue of feature selection. In the second step, feature engineering is performed manually to tailor the approach to the specific signal, which can be constrained the performance of SA detection. Recently, deep learning (DL) has revolutionized the landscape of SA detection by enabling automatic and robust feature extraction, which significantly overcome the drawback of conventional machine learning method above. 

\textbf{Deep learning based methods}: Wang \textit{et al.} \cite{wang2019} introduced a modified LeNet-5 convolutional neural network that incorporates an adjacent segment from a single lead signal. Chen \textit{et al.} \cite{chen2021} engineered the SE-MSCNN, a multi-scaled fusion network specifically designed for SA detection via single-lead ECGs. Yang \textit{et al.} \cite{yang2022} then took this a step further by unveiling a one-dimensional squeeze-and-excitation (SE) residual group network that intricately captures the interplay between heart rate variability (HRV) and ECG-derived respiration (EDR). In another works, Chen \textit{et al.} \cite{chen2023} proposed the BAFNet. This innovative architecture employs a bottleneck attention-based fusion network, targeting key ECG parameters like R-R intervals and R-peak amplitudes, and synergizing convolutional networks with a global query method for a more precise identification of SA. 

However, the feature extraction in these deep learning methods has a disadvantage as it only utilizes the R peaks and RR intervals as input information for the CNN model. As mentioned earlier, relying exclusively on R-peaks and RR intervals could lead to the oversight of critical classification features from other peaks.

Therefore, the paper proposes a solution called MPCNN to address both the problems associated with traditional machine learning-based approaches and some existing deep learning-based approaches. This solution incorporates deep learning to automate the classification process, thereby overcoming the limitations of traditional machine learning methods. Furthermore, by utilizing MP algorithms to calculate the distance profile within each subsequence of the signal, the proposed method captures more information from the input signal. This addresses the drawbacks of certain existing deep learning-based methods and enhances the performance of SA detection.

\subsection{Matrix Profile algorithms}
MP algorithms make contributions to time series problems by identifying information related to discord and motif points, recurrent patterns or subsequences in a time series dataset, and the most exceptional or rare patterns in a time series problem \cite{vchandola2009, amueen2009}. For instance, Silva \textit{et al.} \cite{silva2019} improved on their previous SiMPle representation for music data mining with SiMPle-Fast, a quicker and more precise tool that specializes in subsequence similarity joins and excels at tasks like cover music detection and thumbnailing.  Similarly, Senobari \textit{et al.} \cite{senobari2018} investigated into the Parkfield event's previously missed foreshocks, discovering seismic event clusters that were active around the time of the mainshock, with the goal of completely decoding their temporal patterns and link to the mainshock. Shi \textit{et al.} \cite{shi2019} took this a step further, devising a novel matrix profile-based technique for rapidly and scalably recognizing and categorizing events in synchrophasor data, producing superior results in real-world PMU case studies. Inspired by these methodologies, this study introduces a novel feature extraction approach based on MP algorithms to address the problem of SA detection from single-lead ECG signals. 

\begin{figure*}[htbp]
    \centering
    \includegraphics[width=0.75\textwidth]{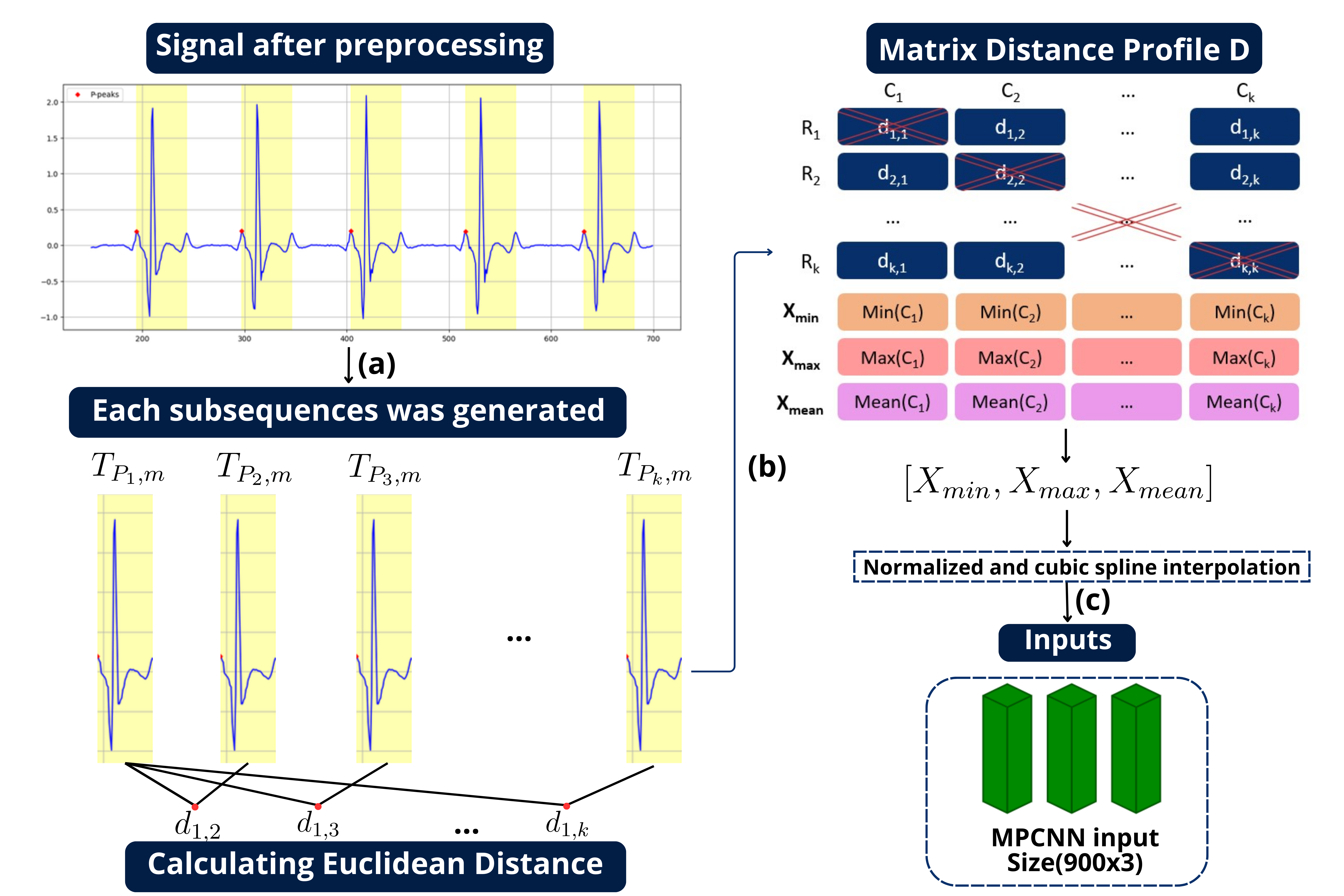}
 \caption{Detailed procedure of the MPCNN. (a) In the preprocessing stage, subsequences $T_{P_i,m}$ are generated, represented by the yellow window in the image. Each subsequence begins at a P peak and extends for a window size of $m$. (b) The Matrix Distance Profile $D$ is created based on the Euclidean Distance, with elements defined as $D_{i,j} = \text{dist}(T_{P_i,m}, T_{P_j,m}), 1 \leq i, j \leq k$. Subsequently, the values of $X_{min}, X_{max}, X_{mean}$ are calculated based on the minimum, maximum, and mean values from each columns of the matrix, excluding the diagonal elements (i.e., $D_{i,i}, 1 \leq i \leq k$). (c) The input is then normalized to the range $[0,1]$ and subjected to cubic spline interpolation to achieve a final dimension of $900 \times 3$.}
    \label{fig:my_label}
\end{figure*}
\section{Methods}

This study proposes a novel method, namely MPCNN, for CNN-based SA classification. The following section provides a detailed description of MPCNN. The discussion initially focuses on preprocessing the initial input signal. Then, the method for constructing new feature extraction using MP algorithms is described. Lastly, the context mentions existing lightweight models used to test the experiment's performance in SA detection.
\subsection{Data preprocessing}
ECG signals often contain various types of noise, such as baseline drift (or baseline wander noise), motion artifacts, and power-line interference\cite{Friesen1990}. To enhance the machine learning model's performance, we undertake specific steps to eliminate noise, as below.
\begin{itemize}
    \item \textbf{Filtering} During this stage, we utilize a Bandpass Finite Impulse Response (FIR) filter, following a similar approach as presented in previous works \cite{wang2019, chen2023}. This selection enables us to isolate significant signal features while eliminating frequencies beyond this range, thereby enhancing the signal-to-noise ratio (SNR) and optimizing the data for subsequent machine learning analysis. 
    \item \textbf{Segmentation}: Some studies \cite{yadollahi2009, maier2000} have demonstrated the effectiveness of incorporating adjacent segment information for efficient SA detection. This approach involves extracting a total of five 1-minute segments, hence we we adopt this method to segment the original signal.
    \item \textbf{Noise removal}: The human maximal heart rate (MHR) is approximately 200 beats per minute \cite{franckowiak2011}. Utilizing this value, along with the minimum rate of around 20 beats per minute, we exclude the segments containing noise from the signal analysis.
\end{itemize}
\subsection{Matrix-Profile-based processing}

To illustrate how MP algorithms could be created new feature extraction for machine learning models in apnea detection. We begin to define some concepts related to this technique. In the proposed approach, we generally adhere to the framework delineated in the works \cite{ChinChiaMichaelYeh, yan2016}. To ensure a comprehensive understanding, we will also provide a review of all essential definitions related to this framework. 

After the pre-processing process, we begin the analysis with a time series \( T \) which is a sequence of real-valued numbers obtained from an ECG signal

\begin{equation}
T = [t_1, t_2, \ldots, t_n], \quad t_i \in \mathbb{R}, \quad 1 \leq i \leq n,
\end{equation}
where \( n \) denotes the length of \( T \).

We define a subsequence \( T_{i,m} \) of the time series \( T \) as a consecutive subset of numbers of length \( m \), starting from position \( i \)
\begin{equation}
T_{i,m} = [t_i, t_{i+1}, \ldots, t_{i+m-1}], \quad 1 \leq i \leq n - m + 1.
\end{equation}

Utilizing the Hamilton algorithm \cite{hamilton2002}, we were able to determine the positions of the R peaks. Based on the positions of these R peaks, we determined the positions of the P peaks, as shown in Algorithm 1. Assuming that \( k \) P peaks have been identified in \( T \), we denote the position of the \( i \)-th P peak as \( P_i \), where \( 1 \leq i \leq k \).

Subsequently, we construct a matrix \( A \) comprising subsequences of \( T \) that start at these P peaks
\begin{equation}
A = [T_{P_1,m}, T_{P_2,m}, \ldots, T_{P_k,m}], \quad T_{P_i,m} \in \mathbb{R}^{1 \times m}.
\end{equation}

From the matrix \( A \), we define a distance profile matrix \( D \) for the ECG signal. The elements of \( D \) are given by
\begin{equation}
D_{i,j} = \text{dist}(T_{P_i,m}, T_{P_j,m}), \quad 1 \leq i, j \leq k,
\end{equation}
where \(\text{dist}(X, Y) = \|X - Y\|\) denotes the Euclidean distance between two subsequences \( X \) and \( Y \) of the same length. These matrices represents the relationship of the distance profiles between each pair of subsequences.

Finally, we derive three input segments from the distance profile matrix \( D \) by considering the minimum, maximum, and mean values in each column, excluding the diagonal elements \( D_{i,i} \) for all \( 1 \leq i \leq k \). Specifically, we calculate

\begin{equation}
X_{\text{min},j} = \min \{D_{i,j} \mid 1 \leq i \leq k, i \neq j\},
\end{equation}
\begin{equation}
X_{\text{mean},j} = \frac{1}{k-1} \sum_{i=1, i \neq j}^{k} D_{i,j},
\end{equation}
\begin{equation}
X_{\text{max},j} = \max \{D_{i,j} \mid 1 \leq i \leq k, i \neq j\},
\end{equation}

for \( 1 \leq j \leq k \), resulting in three vectors \( X_{\text{min}} \), \( X_{\text{max}} \), and \( X_{\text{mean}} \) of length \( k \). We denote these vectors as MinDP, MaxDP, and MeanDP, respectively.

Finally, the input to the model, represented as $[X_{\text{min}}, X_{\text{max}}, X_{\text{mean}}]$, was normalized to fall within the range $[0, 1]$. Additionally, cubic spline interpolation was applied to increase the data size to $900 \times 3$, which allowing the model to learn and generalize better from the input data. The core idea of MPCNN is to extract as much information as possible about the overall distance relationships within the ECG signal. This is achieved by modifying certain parts of the original MPCNN to make it suitable for ECG signal data and to serve as an input for the CNN model. The visualization of this process was shown in Fig. 2.

% \begin{algorithm}
% \caption{Algorithm for determining the ECG P peaks}
% \textbf{Input:} ecg\_signal $E$, r\_peaks $R$, search window $sw = 20$, exclude window $ew = 5$ \\
% \textbf{Output:} List of p\_peaks $P$

% \begin{algorithmic}[1]
% \Procedure{FindPPeaks}{$E, R, sw, ew$}
%     \State $P \gets$ list() \Comment{List for P peaks}
%     \State $N \gets$ length($R$) \Comment{Length of R peaks}
%     \For{$i \gets 0$ \textbf{to} $N-1$}
%         \State $r \gets R[i]$ \Comment{Current R peak}
%         \State $s\_start \gets$ max(0, $r - sw$) \Comment{Search start}
%         \State $s\_end \gets$ max(0, $r - ew$) \Comment{Search end}
%         \If{$s\_start < s\_end$}
%             \State $p\_rel \gets$ index of max($E[s\_start:s\_end]$) \Comment{Relative P peak}
%             \State $p\_abs \gets p\_rel + s\_start$ \Comment{Absolute P peak}
%             \State $P$.append($p\_abs$)
%         \EndIf
%     \EndFor
%     \State \Return $P$
% \EndProcedure
% \end{algorithmic}
% \end{algorithm}

\begin{algorithm}
\caption{\textbf{Algorithm for determining the ECG P peaks}}
\textbf{Input:} A time series ECG signal $E$, list of R peaks $R$, first window $w_1 = 20$, second window $w_2 = 5$ \\
\textbf{Output:} List of P peaks index $P_{\text{index}}$ 
\begin{algorithmic}[1]
\Procedure{FindPPeaks}{$E$, $R$, $w_1$, $w_2$}
    \State $P_{\text{index}} \gets$ list()
    \State $N \gets$ length($R$)
    \For{$i \gets 0$ \textbf{to} $N-1$}
        \State $r \gets R[i]$
        \State $d_1 \gets$ max(0, $r - w_1$)
        \State $d_2 \gets$ max(0, $r - w_2$)
        \If{$d_1 < d_2$}
            \State $\text{index} \gets \text{np.argmax}(E[d_1:d_2])$
            \State $P_{\text{index}}$.append($\text{index} + d_1$)
        \EndIf
    \EndFor
    \State \Return $P_{\text{index}}$
\EndProcedure 
\end{algorithmic}
\end{algorithm}

\subsection{Architecture of CNN model}
\subsubsection{LeNet-5}
The standard LeNet-5 aims to solve character recognition was proposed by LeCun\cite{lecun2015}. The primary objective of this work was to adapt the LeNet-5 architecture to accommodate the unique challenges associated with time series analysis, particularly when processing one-dimensional data. Unlike image or character recognition tasks that predominantly handle two-dimensional data, the dataset in this study presents one-dimensional temporal data. Additionally, due to its smaller size compared to typical datasets used in character or image classification tasks, there's a heightened risk of overfitting. To address these issues and adapt the architecture to handle SA detection in a binary classification problem, we made several modifications to modified LeNet-5 as below:
\begin{itemize}
    \item The shift was made from the two-dimensional convolution operation to a one-dimensional convolution operation, provided a more appropriate choice for feature extraction from time series data \cite{kiranyaz2016}.
    \item Dropout layers were included between the convolution layers and fully connected layers to prevent over-fitting \cite{srivastava2014}.
    \item For the purpose of achieving a balance between network complexity and performance, only a single fully connected layer was retained, in contrast to the multiple layers found in many deep learning models \cite{mam2018}.
    \item A dropout layer with a rate of 0.5 was introduced between the convolution layers and the fully connected layer, which is different from the 0.8 rate in the prior modified model.
    \item For the specific task of binary classification, our method reduced the number of output nodes in the adapted version of the LeNet-5 model from ten to only two.
\end{itemize}

\begin{table*}
\centering
\caption{DETAILS OF MODIFIED LENET-5 CONVOLUTIONAL NEURAL NETWORK.}
\small
\begin{tabular}{|l|l|l|l|}
\hline
\textbf{Layer}     & \textbf{Parameter}           & \textbf{Output Shape} & \textbf{Number} \\ \hline
Input              & -                            & (None, 900, 2)        & 0                             \\ \hline
Conv1              & 64 x 5 x 2, stride 2, pad 0  & (None, 448, 64)       & 832                           \\ \hline
BatchNorm1         & -                            & (None, 448, 64)       & -                             \\ \hline
MaxPooling1        & 3, stride 3, pad 0           & (None, 149, 64)       & 0                             \\ \hline
Dropout1           & 0.5 rate                     & (None, 149, 64)       & 0                             \\ \hline
Conv2              & 96 x 5 x 2, stride 2, pad 0  & (None, 73, 96)        & 12336                         \\ \hline
BatchNorm2         & -                            & (None, 73, 96)        & -                             \\ \hline
MaxPooling2        & 3, stride 3, pad 0           & (None, 24, 96)        & 0                             \\ \hline
Dropout2           & 0.5 rate                     & (None, 24, 96)        & 0                             \\ \hline
Conv3              & 128 x 5 x 2, stride 2, pad 0 & (None, 10, 128)       & 24640                         \\ \hline
BatchNorm3         & -                            & (None, 10, 128)       & -                             \\ \hline
GlobalMaxPooling1D & -                            & (None, 128)           & 0                             \\ \hline
Dropout3           & 0.5 rate                     & (None, 128)           & 0                             \\ \hline
FC1                & 128, relu                    & (None, 128)           & 16512                         \\ \hline
Dropout4           & 0.5 rate                     & (None, 128)           & 0                             \\ \hline
FC2                & 64, relu                     & (None, 64)            & 8256                          \\ \hline
Dropout5           & 0.5 rate                     & (None, 64)            & 0                             \\ \hline
Output             & 2, softmax                   & (None, 2)             & 130                           \\ \hline
\end{tabular}
\end{table*}

The architectural design and detailed specifications of the proposed modified LeNet-5 are comprehensively presented in Table I.
\subsubsection{BAFNet and SE-MSCNN} 
To evaluate the proposed approach's effectiveness, its performance was tested alongside existing models used in the SA detection problem. These models, namely BAFNet and SE-MSCNN, introduced by Chen \textit{et al.} \cite{chen2021} and Chen \textit{et al.} \cite{chen2023}, are lightweight, with total sizes of only 0.29MB and 0.16MB, respectively. These compact models are well-suited for real-world applications, including HSAT or SA detection on IoT devices.
\begin{itemize}

    \item BAFNet is a deep learning model that has been optimized for time-series data analysis, specifically designed for detecting SA in single-lead ECG signals. The architecture of BAFNet consists of five main components: the RRI stream network, the RPA stream network, global query generation, feature fusion, and a classifier. The RRI and RPA stream networks are designed based on encoder and decoder layers to effectively extract important features from R peaks and RR intervals. In this study, the RRI and RPA stream networks utilize alternate values of MinDP and MaxDP, respectively. The global query generation component is included in the architecture to efficiently extract highly relevant information from the input while minimizing computational costs. Lastly, the feature fusion component is tasked with learning the significance of feature maps, which are flexibly extracted from various stream networks. Additionally, the author introduces a hard sample mining method used in the second stage, following the initial stage that employs solely a BAFNet model. However, the proposed approach MPCNN utilizes only the BAFNet model from the first stage to test with new feature extraction.
    
    \item SE-MSCNN is another lightweight model designed for SA classification. The model comprises two main components: a Multi-scaled CNN module and a channel-wise attention module. The first component involves the use of different time scales for adjacent segments, specifically at 1-minute, 3-minute, and 5-minute intervals, based on the original signal. These time-scaled segments then serve as inputs for the CNN model. The second component employs a channel-wise attention mechanism to adapt to the system of adjacent segments. In the original research, these adjacent segments contained information about R peaks and RR intervals. However, in the proposed approach, we use MinDP and MaxDP values, similar to those in the previous experiment with BAFNet. We also divide the feature extraction into adjacent segments of 1-minute, 3-minute, and 5-minute intervals, retaining all the architectural elements originally proposed by the SE-MSCNN method.
\end{itemize}

\section{Experiments}

\subsection{Datasets and experimental settings}
The dataset employed for the experiments is sourced from the PhysioNet Apnea-ECG repository, which is widely recognized for its comprehensive collection of ECG recordings
 \cite{penzel2000, goldberger2000}. A total of 70 overnight recordings were obtained from a sample of 70 participants, consisting of 57 males and 13 females. The participants were divided into two groups, with half of them assigned to the training set and the remaining participants assigned to the test set prior to data collection. The dataset contains ECG signals with a sampling frequency of 100 Hz. The duration of each ECG signal ranges from little under 7 hours to approximately 10 hours. Based on the concurrent signals of respiration and oxygen saturation, a group of experts formulated a series of annotations for segments lasting one minute, distinguishing between periods with apnea (A) and periods without apnea (N). Table II presents the comprehensive details of the PhysioNet Apnea-ECG database. Furthermore, the division of the training set and test set is conducted at the recording level, ensuring that each nightly recording is exclusively included in either the training set or the test set, but not both.

\begin{table}[H]
\centering
\caption{ DESCRIPTION OF THE PHYSIONET APNEA-ECG DATASET.}
\begin{adjustbox}{width=0.5\textwidth}
\begin{tabular}{lccc}
\hline
                          & \multicolumn{1}{l}{Training set} & \multicolumn{1}{l}{Test set} & \multicolumn{1}{l}{Total} \\ \hline
Number of participants        & 35                               & 35                           & 70                        \\
Number of male participants   & 30                               & 27                           & 57                        \\
Number of female participants & 5                                & 8                            & 13                        \\
Number of patient         & 23                               & 23                           & 46                        \\
Patient ratio             & 65.71\%                          & 65.71\%                      & 65.71\%                   \\ \hline
\end{tabular}
\end{adjustbox}
\end{table}

\begin{table}[H]
\centering
\caption{DESCRIPTION OF THE SEGMENTS IN PHYSIONET APNEA-ECG DATASET.}
\begin{adjustbox}{width=0.5\textwidth}
\begin{tabular}{cccc}
\hline
              & Training Set   & Validation Set & Test Set        \\ \hline
SA             & 5180 (38.74\%) & 1296 (38.74\%) & 5491 (38.30\%)  \\
Non SA         & 8190 (61.26\%) & 2048 (61.26\%) & 10455 (61.70\%) \\
\textbf{Total} & \textbf{13,370} & \textbf{3,343}  & \textbf{16,946}  \\ \hline
\end{tabular}
\end{adjustbox}
\end{table}

\subsection{Model training and testing}
\subsubsection{Per-segment analysis}
After completing the preprocessing stage, a total of 16,713 segments were extracted from the released set (35 subjects), and 16,946 segments from the withheld set (35 subjects) within the Apnea-ECG dataset. To prepare for model training, the data were divided into different sets. The released set was designated for training, and the withheld set was reserved for testing. Additionally, the training set was further partitioned into training and validation subsets at a ratio of 70\%:30\%. A detailed depiction of the data distribution can be found in Table III. 

The model's hyperparameters were selected to optimize the performance. The training process was conducted over 100 epochs, using a batch size of 128 and an initial learning rate of 0.001. To ensure efficient convergence, the learning rate was kept constant until the 70th epoch. Thereafter, it was gradually reduced by 10\% every ten epochs. The optimization process employed the Adaptive Moment Estimation (Adam) algorithm \cite{kingma2014}, which is known for its effectiveness in classification problems. The experiments in this study were conducted on the Google Colab platform, which was configured with a Python 3 runtime environment. A Tesla T4 GPU equipped with high RAM was utilized as the hardware accelerator. TensorFlow and Keras libraries, both at version 2.12.0, were used for the development and execution of the machine learning models evaluated in this research.

In the case of BAFNet, only the model's first stage is utilized. This decision is based on the fact that the characteristics associated with the second stage, particularly hard samples, depend on the specifics of the dataset. For example, the value of $P$ could be useful in hard samples because it shows the significance between each component in the input. However, values like MinDP, MeanDP, and MaxDP do not possess these properties. Moreover, determining an optimal clustering value for a new feature requires experimental investigation, which is not the focus of this paper.

\subsubsection{Per-recording analysis}
After the model has been trained using the released set, its performance was evaluated on a per-recording basis using only the Apnea-ECG withheld set, which consists of 35 recordings that were kept blind to the model. This evaluation approach serves a critical purpose in testing how well the model generalizes to unseen data. By employing a withheld set that the model has not been exposed to during the training phase, we are able to assess its ability to accurately classify new, unseen instances. This provides a more robust and realistic understanding of the model's effectiveness and reliability in real-world applications.
\subsection{Performance metrics}
\subsubsection{Per-segment metrics}
In the per-segment evaluation, we employ several widely recognized metrics to measure the performance of the model. Each of these metrics provides a unique perspective on the accuracy and robustness of the model in classifying the segments. The key metrics used are:

\begin{itemize}
    \item \textbf{Accuracy (Acc):} This metric provides a general measure of how well the model has classified both positive and negative instances. It is defined as:
    \[
    \text{Acc} = \frac{TP + TN}{TP + TN + FP + FN}
    \]
    \item \textbf{Specificity (Spec):} Specificity measures the proportion of actual negatives that are correctly identified. It is calculated as:
    \[
    \text{Spec} = \frac{TN}{TN + FP}
    \]

    \item \textbf{Sensitivity (Sen):} Also known as the true positive rate, sensitivity quantifies the proportion of actual positives that are correctly identified. It is defined as:
    \[
    \text{Sen} = \frac{TP}{TP + FN}
    \]

    \item \textbf{\(F_1\) Score:} The \(F_1\) score is the harmonic mean of precision and recall, providing a balance between these two metrics. It is calculated as:
    \[
    F_1 = \frac{2 \times TP}{2 \times TP + FP + FN}
    \]
    where TP, TN, FP, and FN are the counts of true positives, true negatives, false positives, and false negatives, respectively.

\end{itemize}

These metrics together provide a comprehensive overview of the model's performance, allowing us to assess its strengths and weaknesses in different aspects of classification.

\subsubsection{Per-recording metrics}
In the per-recording evaluation, we assess the model's performance by considering metrics at the recording level. This perspective helps in understanding the model's ability to predict and categorize entire recordings rather than individual segments. In addition to accuracy, sensitivity, and specificity, as defined earlier, we introduce:

\begin{itemize}
    \item \textbf{Pearson Correlation Coefficient (Corr):} This metric measures the linear relationship between predicted and actual values. It is an essential measure to evaluate the degree of correspondence between the model's predictions and actual observations.

    \item \textbf{Apnea-Hypopnea Index (AHI):} AHI is an essential criterion used to determine whether a subject has normal breathing or apnea. It's computed as:
\[
\text{AHI} = \frac{60 \times T}{N},
\]
where \( T \) represents the entire count of one-minute intervals observed during the overnight recording, and \( N \) corresponds to the count of one-minute intervals that include SA segments. A subject with an AHI value below 5 is considered normal, while a value of 5 or greater indicates apnea \cite{ruehland2009}.
\end{itemize}
These per-recording metrics, coupled with the per-segment
metrics, provide a multi-faceted evaluation of the model’s performance, reflecting both its precision in classifying individual
segments and its effectiveness in predicting the overall state
of subjects.

\subsection{Classification performance}
The proposed method for new feature extraction is evaluated using the PhysioNet Apnea-ECG dataset. Detailed performances for both per-segment and per-recording classifications are presented below:

\subsubsection{Per-segment performance}
As illustrated in Fig. 3(a), Fig. 4(a), and Fig. 4(b), our novel feature extraction technique based on distance profiles achieved in main experiment modified LeNet-5 accuracy of \(91.89\%\), sensitivity of \(89.36\%\), specificity of \(93.46\%\), and F0 of \(89.41\%\). Fig. 5(a) displays the confusion matrix for performance in the modified LeNet-5 model, demonstrating that the majority of segments were correctly classified. To evaluate the robustness of our new feature extraction, we tested our approach on two different pre-existing lightweight models without altering their architectures. In both cases—SE-MSCNN and BAFNet - the proposed method outperformed the traditional R-peaks and RR-interval approaches. Detailed comparisons are presented in Table V. The BAFNet model (Stage 1) achieved an accuracy of \(91.83\%\), sensitivity of \(87.67\%\), specificity of \(94.41\%\), and F0 of \(89.16\%\). Likewise, for SE-MSCNN, the performance metrics reached \(92.11\%\) accuracy, \(87.73\%\) sensitivity, \(94.83\%\) specificity, and \(89.5\%\) F0.

\begin{figure}[H]
    \centering
    \begin{subfigure}[b]{0.48\linewidth}
        \includegraphics[width=\linewidth]{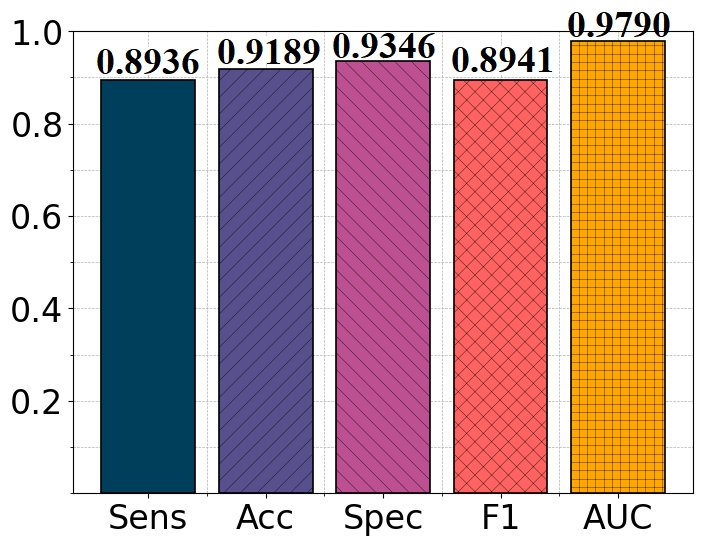}
        \subcaption{}
        \label{fig:persegment}
    \end{subfigure}
    \hfill
    \begin{subfigure}[b]{0.48\linewidth}
        \includegraphics[width=\linewidth]{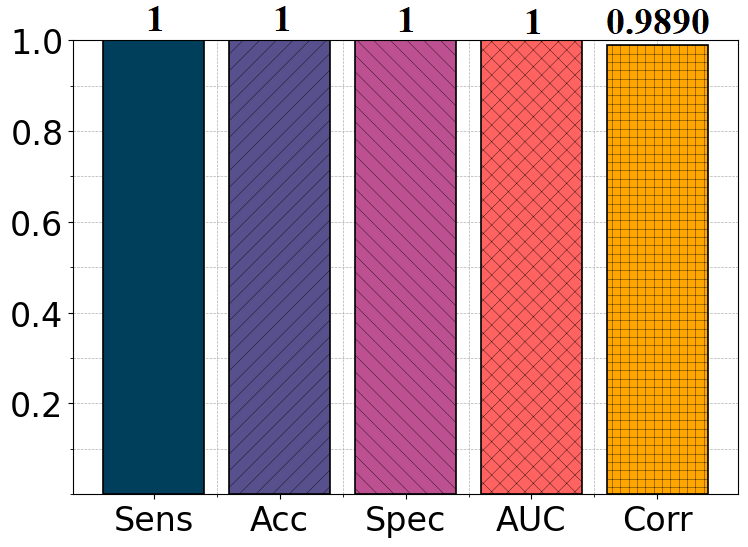}
        \subcaption{}
        \label{fig:perrecording}
    \end{subfigure}
    \caption{The bar charts of main CNN(LeNet-5) in SA detection performance on PhysioNet Apnea-ECG dataset: (a) per-segment; (b) per-recording.}
    \label{fig:common}
\end{figure}
\begin{figure}[H]
    \centering
    \begin{subfigure}[b]{0.48\linewidth}
        \includegraphics[width=\linewidth]{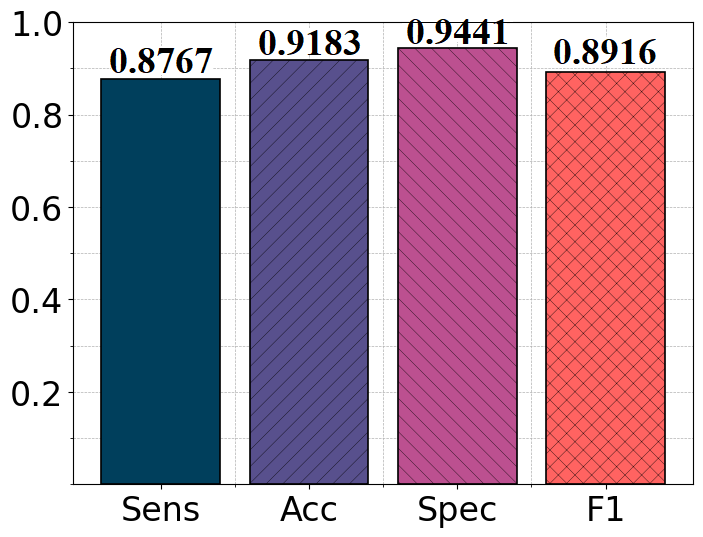}
        \subcaption{}
        \label{fig:persegment}
    \end{subfigure}
    \hfill
    \begin{subfigure}[b]{0.48\linewidth}
        \includegraphics[width=\linewidth]{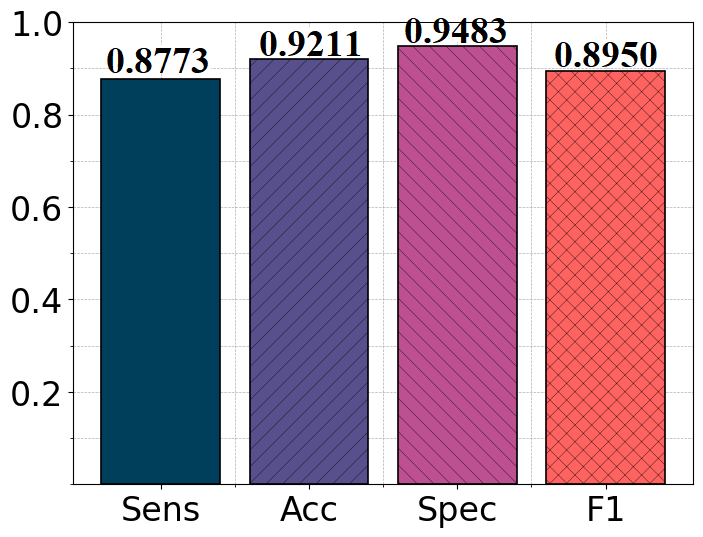}
        \subcaption{}
        \label{fig:perrecording}
    \end{subfigure}
    \caption{The bar charts of performance in per-segment from two additional CNNs: (a) BAFNet; (b) SE-MSCNN.}
    \label{fig:common}
\end{figure}
\begin{figure}[H]
    \centering
    \begin{subfigure}[b]{0.48\linewidth}
        \includegraphics[width=\linewidth]{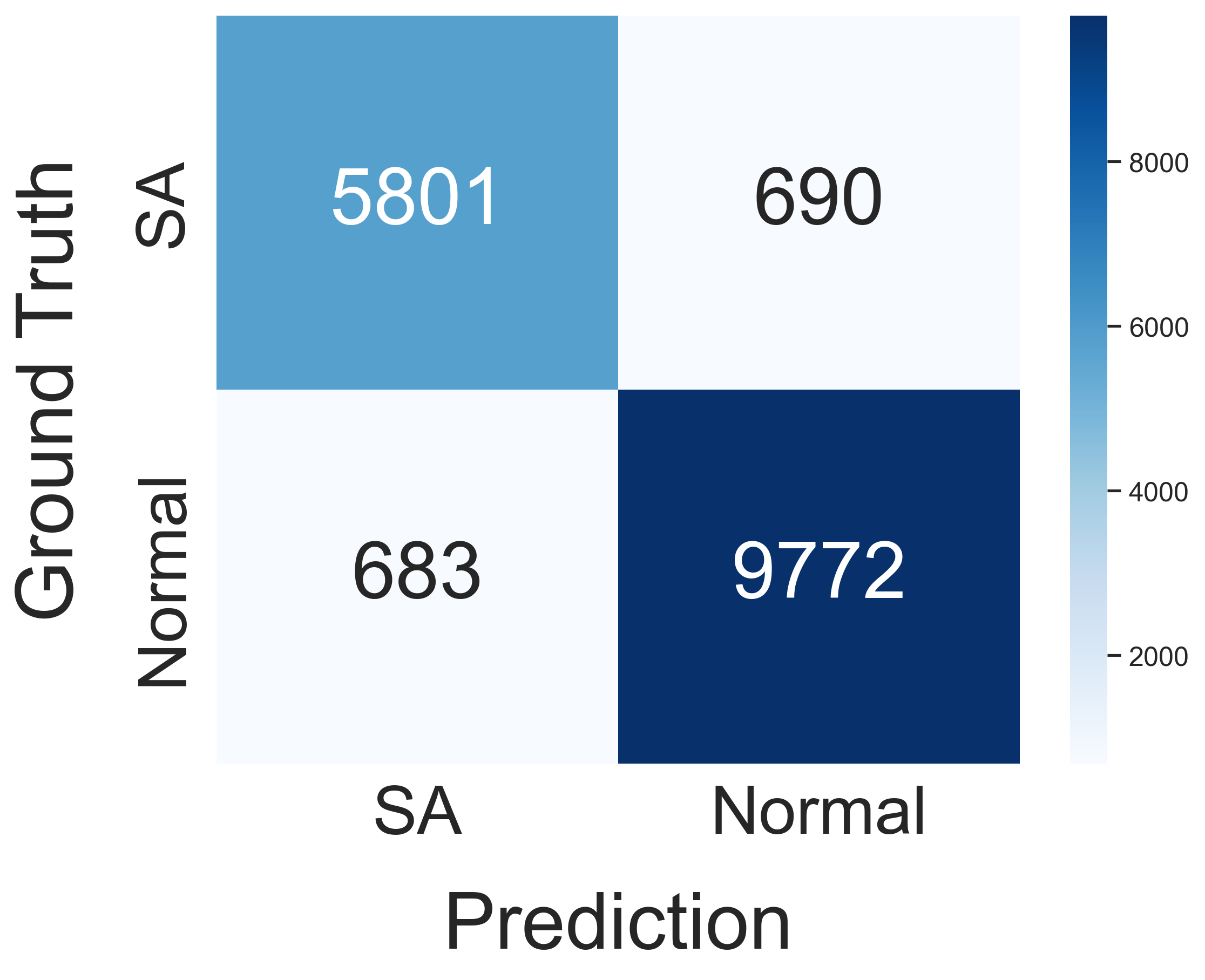}
        \subcaption{}
        \label{fig:persegment}
    \end{subfigure}
    \hfill
    \begin{subfigure}[b]{0.48\linewidth}
        \includegraphics[width=\linewidth]{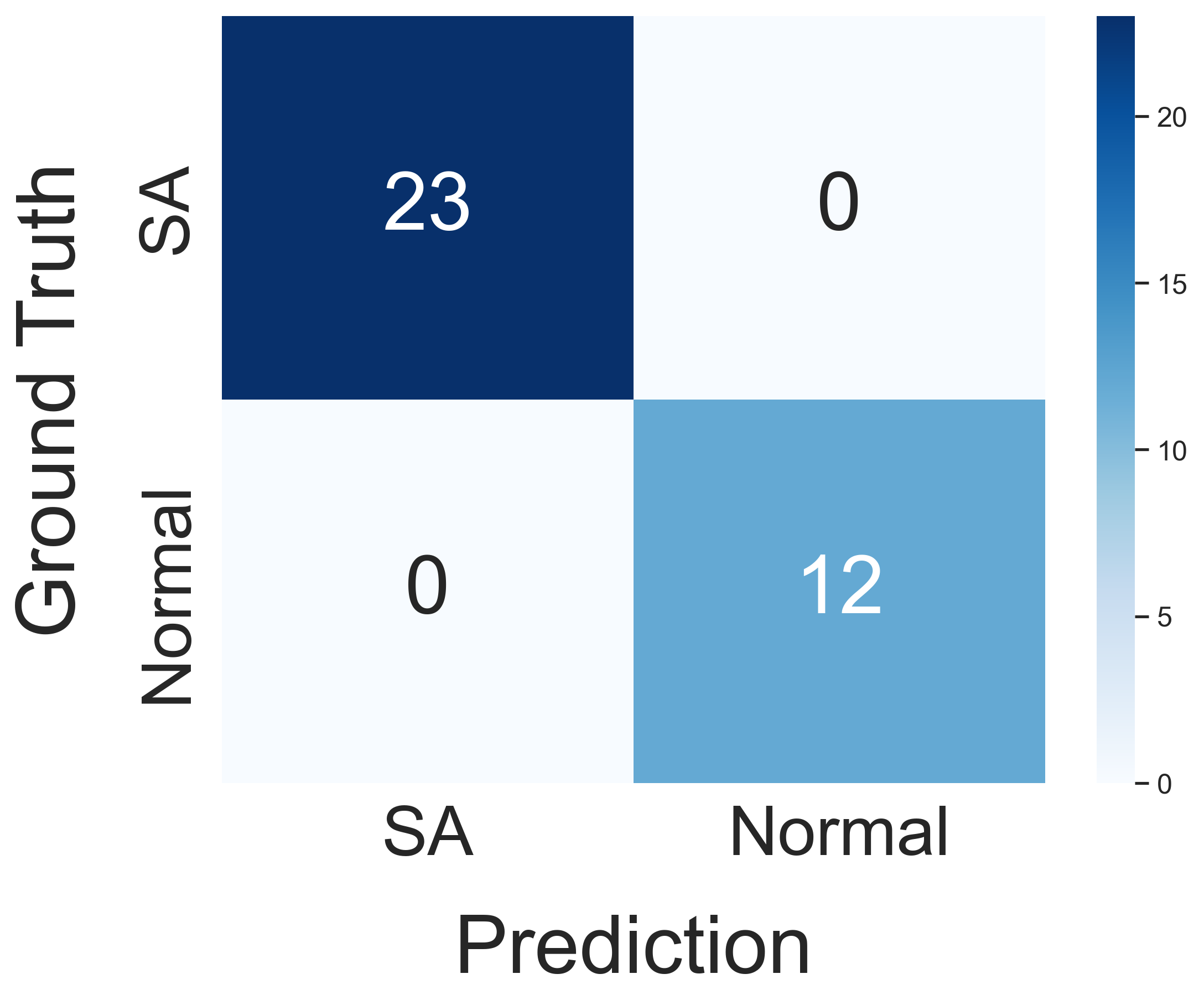}
        \subcaption{}
        \label{fig:perrecording}
    \end{subfigure}
    \caption{Confusion matrix on PhysioNet Apnea-ECG dataset in modified LeNet-5 models: (a) per-segment; (b) per-recording.}
    \label{fig:common}
\end{figure}

\subsubsection{Per-recording performance}
According to Fig. 3(b), the primary CNN model used in this study, LeNet-5, achieved full scores across all metrics—accuracy, sensitivity, specificity, and AUC value—all at \(100\%\). Furthermore, a high correlation value of \(0.989\) was attained, indicating strong agreement between the ground truth Apnea-Hypopnea Index (AHI) and the predicted AHI from overnight ECG recordings. Fig. 5(b) shows the confusion matrix performance for SA detection in per-recording performance.

\subsection{Ablation study}
In order to verify the effectiveness of each component in MPCNN, we conducted two main experiments. The first experiment was designed to evaluate the effectiveness of each component: MinDP, MaxDP, and MeanDP. This experiment was use model $T_1$ to test. The second experiment aimed to assess the effectiveness of different window sizes, represented by the value of $m$. The best criteria in first experiment was utilized in the second experiment. Each experiment was run 5 times in modifed LeNet-5. In the first experiment, there were a total of seven combinations of three values: MinDP, MaxDP, and MeanDP, while the second experiment featured a total of four cases, each designed to capture different information from each segment. Table IV and Fig. 6. was shown detail about each experiments.
\begin{table}[H]
\centering
\caption{THE DESIGN OF THE FIRST ABLATION STUDY.}
\begin{tabular}{|c|c|c|c|}
\hline
Methods & MinDP & MaxDP & MeanDP\\ 
\hline
M1 & \checkmark & &  \\
\hline
M2 & & \checkmark &  \\
\hline
M3 &  &  & \checkmark \\
\hline
M4 & \checkmark & \checkmark &   \\
\hline
M5 &  & \checkmark & \checkmark  \\
\hline
M6 & \checkmark &  & \checkmark   \\
\hline
M7 & \checkmark & \checkmark & \checkmark   \\
\hline
\end{tabular}
\end{table}

\begin{table*}
\centering
\caption{COMPARISON WITH THE CONVENTIONAL FEATURE EXTRACTION.}
\begin{adjustbox}{width=1\textwidth}
\begin{tabular}{llclclclclclc}
\hline
                                                         &  & \multicolumn{5}{c}{BAFNet (Stage 1) \cite{chen2023}}                      & \multicolumn{1}{c}{} & \multicolumn{5}{c}{SE-MSCNN \cite{chen2021}}                                 \\ \cline{3-7} \cline{9-13} 
                                                         &  & Acc         &  & Sens   &  & Spec      &                      & Acc        &  & Sens     &  & Spec      \\ \cline{1-1} \cline{3-3} \cline{5-5} \cline{7-7} \cline{9-9} \cline{11-11} \cline{13-13} 
Original: R peaks and RR interval &  & 90.92\%          &  & \textbf{89.00\%} &  & 92.60\%          &                      & 90.60\%          &  & 86.00\%             &  & 93.50\%          \\
MPCNN: MinDP, MaxDP and MeanDP &  & \textbf{91.83\%} &  & 87.73\%       &  & \textbf{94.41\%} &                      & \textbf{92.11\%} &  & \textbf{87.73\%} &  & \textbf{94.83\%} \\ \hline
\end{tabular}
\end{adjustbox}
\end{table*}

\begin{figure}[H]
    \centering
    \includegraphics[width=0.55\columnwidth]{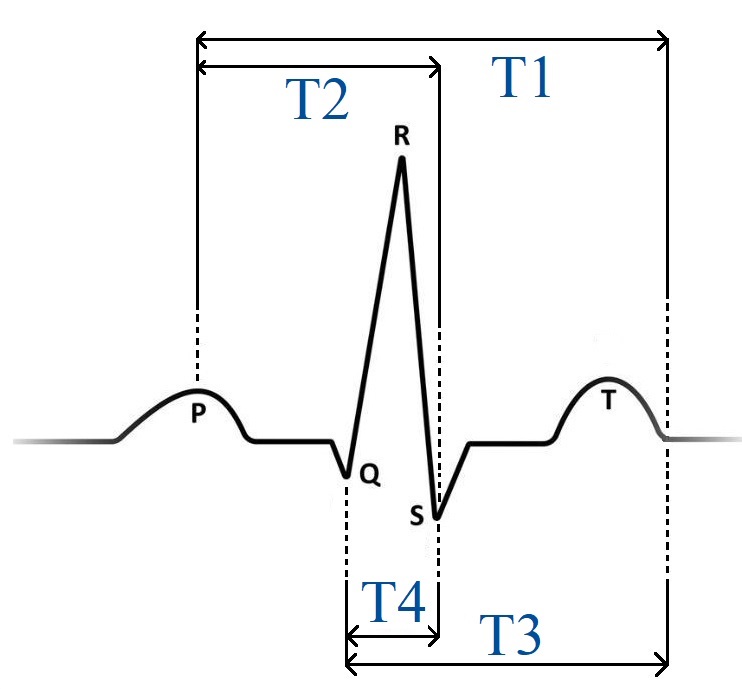}
    \caption{The design of the second ablation experiments with different window size. \\ $T_1$: From P peak to the end of ST segment. \\$T_2$: From P peak to the end of QRS segment. \\$T_3$: Q peaks to the end of ST segment. \\$T_4$: QRS segment.}
\end{figure}

Tables VI and VII present the performance metrics for two distinct ablation experiments. In the first experiment, which evaluates the influence of different input components, models \(M_1, M_4, M_6, M_7\) with MinDP values outperformed models \(M_2, M_3, M_5\), which utilize MaxDP and MeanDP values. 
Specifically, model \(M_1\) achieved an accuracy of \(91.18\% \pm 0.04\%\), and had the highest specificity at \(94.03 \pm 0.3\%\). Interestingly, the inclusion of MaxDP values in model \(M_4\) resulted in a marginal improvement of \(0.58\%\) in accuracy and \(3.05\%\) in sensitivity compared to \(M_1\).
Conversely, while the inclusion of MeanDP values reduced per-segment accuracy in \(M_6\)  , it enhanced the per-recording performance to \(100\%\) across all metrics. and achieved a Corr value of \(0.989\). In model \(M_7\), where all features—MinDP, MaxDP, and MeanDP—were included, the sensitivity was the highest at \(91.34 \pm 0.34\%\). Although this model did not have the highest accuracy, its sensitivity was \(0.39\%\) higher than that of the next best model, \(M_6\).

Besides that, MinDP emerges as the most critical feature in the model. When combined with MaxDP in model \(M_4\), it yields the highest accuracy in terms of per-segment performance. Similarly, the combination of MinDP and MeanDP in model \(M_6\) results in $100 \%$ score across all metrics for per-recording performance, achieving a Corr value of \(0.989\). In the scenario where all combinations of MinDP, MaxDP, and MeanDP are employed in model \(M_7\), the highest sensitivity value of \(91.34 \pm 0.34 \%\) is achieved.

In the second experiment, we focused on model \(M_4\) and varied the window sizes \(T_1, T_2, T_3, T_4\). It became evident that \(T_1\), which has the largest window size extending from the P peak to the end of the ST segment, demonstrated the best performance. The accuracy reached up to \(91.76 \pm 0.13 \%\), exceeding the second-best performance by \(0.84 \%\). Not only did \(T_1\) achieve the highest accuracy, but it also obtained the best sensitivity and specificity scores, reaching \(89.64 \pm 0.31 \%\), and \(93.07 \pm 0.38 \%\) respectively. Overall, the window that encapsulates the complete information in a single segment, including the PQRST segment represented by \(T_1\), yields the best performance in MPCNN.

\begin{table}[H]
\centering
\caption{THE RESULT OF THE SECOND ABLATION STUDY.}
\begin{tabular}{|l|c|c|c|}
\hline
Method & Acc(\%) & Sens(\%) & Spec(\%) \\
\hline
T1 & \textbf{91.76$\pm$0.13} & \textbf{89.64$\pm$0.31} & \textbf{93.07$\pm$0.38} \\
\hline
T2 & 90.94$\pm$0.07 & 87.46$\pm$0.28 & 93.03$\pm$0.26 \\
\hline
T3 & 89.32$\pm$0.04 & 84.70$\pm$0.51 & 92.19$\pm$0.36\\
\hline
T4 & 90.13$\pm$0.09 & 87.64$\pm$0.41 & 91.75$\pm$0.18 \\
\hline
\end{tabular}
\end{table}
\subsection{Performance comparison}

\subsubsection{Per-segment classification} The performance comparisons for per-segment classification against state-of-the-art SA detection methods are presented in TABLE VIII. The results indicate that MPCNN outperforms conventional feature extraction techniques used in prior research. Using the modified LeNet-5 model, we achieved superior results in three key metrics: \(91.89 \%\) accuracy, \(89.36 \%\) sensitivity, and \(93.46 \%\) specificity. These are higher than the results from Wang \textit{et al.} \cite{wang2019} by \(4.29 \%\), \(6.26 \%\), and \(3.16 \%\), respectively.
\begin{table*}[b]
\centering
\caption{THE RESULT OF THE FIRST ABLATION EXPERIMENTS WITH DIFFERENT WINDOW SIZE. }
\begin{tabularx}{\textwidth}{l|XXX|XXXXX}
    \hline
    \multirow{2}{*}{Method} & \multicolumn{3}{c|}{Per-segment} & \multicolumn{5}{c}{Per-recording}  \\
    & Acc (\%) & Sens (\%) & Spec (\%) & Acc (\%) & Sens (\%) & Spec (\%) & AUC & Corr \\
    \hline
    M1 & 91.18$\pm$0.04 & 86.59$\pm$0.59 & \textbf{94.03$\pm$0.30} & 97.14 & 95.65 & 100 & 1 & \textbf{0.991} \\
    M2 & 82.6$\pm$0.18 & 76.09$\pm$0.67 & 86.66$\pm$0.59 & 82.86 & 100 & 50 & 1 & 0.932 \\
    M3 & 84.15$\pm$0.04 & 78.64$\pm$0.13 & 87.57$\pm$0.13 & 88.57 & 100 & 66.67 & 0.985 & 0.963 \\
    M4 & \textbf{91.76$\pm$0.13} & 89.64$\pm$0.31 & 93.08$\pm$0.38 & 97.14 & 95.65 & 100 & 1 & 0.990 \\
    M5 & 85.04$\pm$0.14 & 80.80$\pm$0.51 & 87.67$\pm$0.16 & 88.57 & 100 & 66.67 & 1 & 0.945 \\
    M6 & 89.98$\pm$0.04 & 90.95$\pm$0.24 & 89.38$\pm$0.09 & \textbf{100} & \textbf{100} & \textbf{100} & \textbf{1} & 0.989 \\
    M7 & 90.91$\pm$0.41 & \textbf{91.34$\pm$0.34} & 90.94$\pm$0.45 & 97.14 & 95.65 & 100 & 1 & 0.989 \\
    \hline
\end{tabularx}
\end{table*}

\begin{figure}[H]
    \centering
    \begin{subfigure}[b]{\linewidth}
        \includegraphics[width=\linewidth]{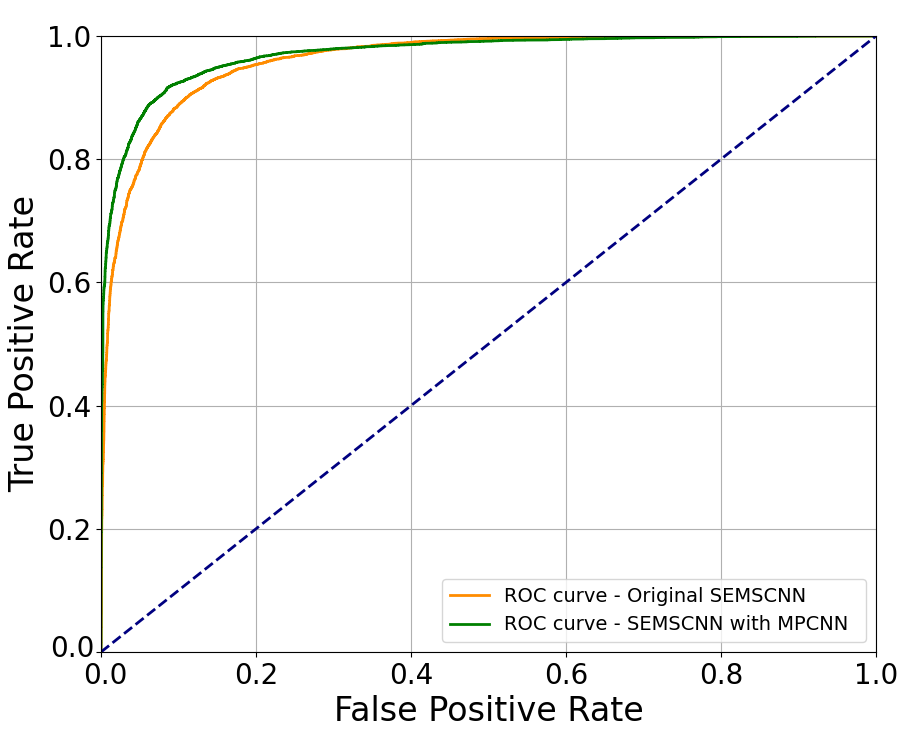}
        \subcaption{}
        \label{fig:rocsescnn}
    \end{subfigure}

    \begin{subfigure}[b]{\linewidth}
        \includegraphics[width=\linewidth]{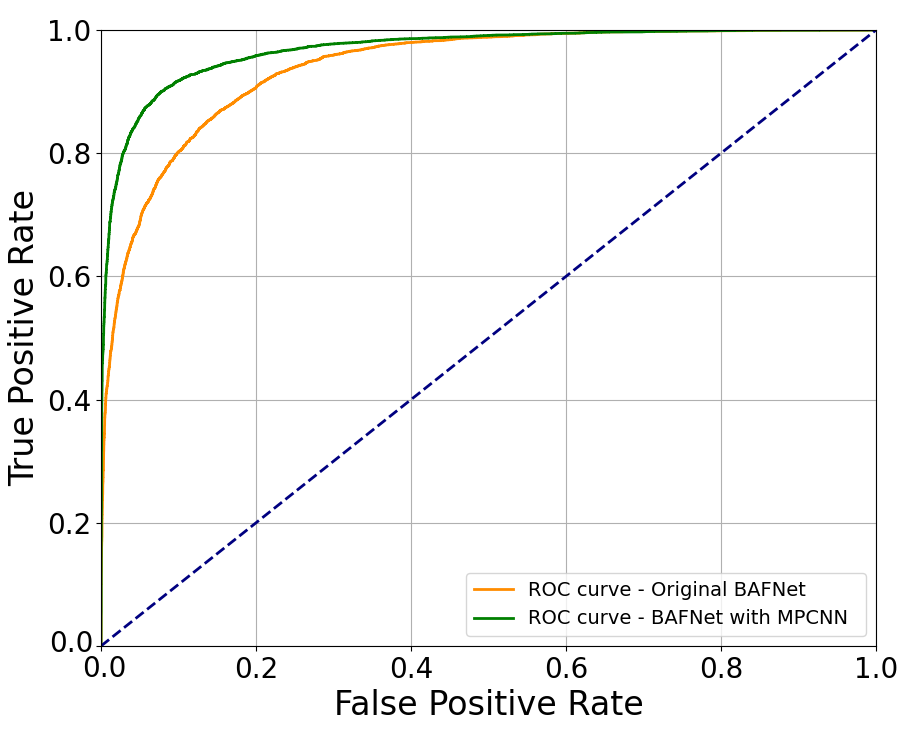}
        \subcaption{}
        \label{fig:rocbafnet}
    \end{subfigure}
    \caption{Comparison of Receiver Operating Characteristic (ROC) Curve with and without our new feature extraction: (a) SEMSCNN; (b) BAFNet}
    \label{fig:common}
\end{figure}
\begin{table*}
\caption{PER-SEGMENT PERFORMANCE COMPARISON WITH STATE-OF-THE-ART METHODS.}
\centering
\resizebox{\textwidth}{!}{%
\begin{tabular}{ccllllll}
\hline
Work                 & Year & \multicolumn{1}{c}{Input Type}                                                         & Feature Extraction  & Classifer       & Acc (\%)      & Sens (\%) & Spec (\%) \\ \hline
Varon \textit{et al.}\cite{varon2015}          & 2015 & \begin{tabular}[c]{@{}l@{}}QRS complex and HRV features\end{tabular}            & Feature Engineering & LS-SVM          & 84.7          & 84.7       & 84.7      \\
Song \textit{et al.}\cite{song2015}          & 2016 & R-R interval and EDR signal                                & Feature Engineering & HMM-SVM         & 86.2          & 82.6       & 88.4      \\
Sharma \textit{et al.}\cite{sharma2016}         & 2016 & \begin{tabular}[c]{@{}l@{}}HRV and Hermite basis       function\end{tabular} & Feature Engineering & LS-SVM          & 83.4          & 79.5       & 88.4      \\
Surrel \textit{et al.}\cite{surrel2018}        & 2018 & RR interval and RS amplitude                                                           & Feature Engineering & SVM             & 85.7          & 81.4       & 88.4      \\
Li \textit{et al.}\cite{li2018}             & 2018 & R-R interval                                                                           & Auto-encoder        & Decision fusion & 84.7          & 88.9       & 82.1      \\
Wang \textit{et al.}\cite{wang2019}           & 2019 & R-R interval and R peaks                                                               & CNN                 & LeNet-5         & 87.6          & 83.1       & 90.3         \\
Sharan \textit{et al.}\cite{sharan2020}        & 2020 & HRV features                                                                           & CNN                 & 1D-CNN          & 88.2          & 82.7       & 91.6      \\
Feng \textit{et al.}\cite{feng2021}           & 2020 & R-R interval                                                                           & Auto-encoder        & TDCS            & 85.1          & 86.2       & 84.4      \\
Bahrami \textit{et al.}\cite{mbahrami2021}        & 2021 & R-R intervals and R-peak amplitude                                                     & CNN                 & LeNet+LSTM      & 80.7          & 75.0         & 84.1      \\
Almutairi \textit{et al.}\cite{almutairi2021} & 2021 & R peaks and R-R intervals                                                              & CNN+LSTM            & LSTM            & 90.9          & \textbf{91.2}       & 90.4      \\
Shen \textit{et al.}\cite{shen2021}         & 2021 & RR Interval                                                                            & MSDA-1DCNN          & WLTD            & 89.4          & 89.8       & 89.1      \\
Chen \textit{et al.}\cite{chen2021}         & 2021 & R peaks and R-R intervals                                                              & CNN                 & SE-MSCNN        & 90.6          & 86.0       & 93.5      \\
Yang \textit{et al.}\cite{yang2022}          & 2022 & HRV and EDR signal                                                                     & CNN                 & 1D-SEResGNet    & 90.3          & 87.6       & 91.9      \\
Bahrami \textit{et al.}\cite{mbahrami2022}       & 2022 & R peaks and R-R intervals                                                              & ZF-Net              & BiLSTM          & 88.1          & 81.5       & 92.3      \\
Chen \textit{et al.}\cite{chen2022}           & 2022 & RR intervals and R-peak amplitudes                                                     & CNN                 & CNN-BiGRU       & 91.2          & 86.4       & 94.1      \\
Chen \textit{et al.}\cite{chen2023}           & 2023 & R peaks and R-R intervals                                                              & CNN                 & BAFNet          & 91.3          & 89.0       & 92.6      \\
Liu \textit{et al.}\cite{liu2023}            & 2023 & Raw ECG signal                                                                         & CNN                 & CNN+Transformer & 88.2          & 78.5       & 94.1      \\
Chen \textit{et al.}\cite{chen2023RAFNet}           & 2023 & R peaks and R-R intervals                                                              & CNN                 & RAFNet          & 91.4          & 88.7      & 93.1     \\
Our[a]                  &   -   & \textbf{MinDP, MaxDP and MeanDP}             & MPCNN              & LeNet-5         & 91.9 & 89.4     & 93.5     \\ 
Our[b]                  &   -   & \textbf{MinDP, MaxDP and MeanDP}             & MPCNN              & BAFNet         & 91.8 & 87.7     & 94.4        \\
Our[c]                  &   -   & \textbf{MinDP, MaxDP and MeanDP}             & MPCNN             & SE-MSCNN         & \textbf{92.1} & 87.7      & \textbf{94.8}     \\\hline
\end{tabular}
}
\end{table*}
However, since we have modified the architecture of LeNet-5, the comparison with the method from Wang \textit{et al.} \cite{wang2019} is not straightforward. To validate the robustness of MPCNN and overcome the problem above, we compared its performance with lightweight models such as BAFNet and SE-MSCNN without altering their architectures. When compared to the architecture BAFNet used by Chen \textit{et al.} \cite{chen2023}, the proposed approach improved performance by \(0.53 \%\) in accuracy and \(1.81 \%\) in specificity. It should be noted that we employed only the first stage of these models without fine-tuning the hard samples technique discussed in section IV-B.1. Even so, MPCNN surpassed the performance of BAFNet in both the first stage and the second stages, which the original one achieved \(90.92 \%\) accuracy in the first stages and improved to \(91.25 \%\) after the second stages. Overall, MPCNN outperforms existing techniques even without using second stages.

Additionally, our results using the same SE-MSCNN architecture as in Chen \textit{et al.} \cite{chen2021} also show improvements of \(1.51 \%\), \(1.73 \%\), and \(1.30 \%\), in terms of accuracy, sensitivity, and specificity, respectively. Furthermore, it's worth noting that all the architectures we employed are lightweight models.  As shown in Fig. 7, the proposed approach’s ROC curve is always above
the original feature extraction in both case BAFNet and SE-MSCNN. These result suggest that the effective and robustness of the new feature extraction with MP Algoritihms.

\subsubsection{Per-recording classification} In the domain of per-recording classification, the modified LeNet-5 model achieved a full score of 100\%, as shown in Table IX. This accomplishment is in harmony with the outcomes that have been previously documented in various studies, specifically, these state-of-the-art methods of \cite{li2018, shen2021, chen2021, chen2023, yang2022}. Notably, while several models have come close to achieving similar precision, MPCNN has an edge that distinguishes it from its peers. One of the most significant aspects that underline its superior performance is the correlation coefficient. MPCNN boasts a correlation coefficient higher than the best-performing baseline model which value of 0.989. This higher correlation coefficient not only underscores the model's accuracy but also indicates its reliability in diverse scenarios, making it a preferred choice for applications requiring robust and consistent results.
\begin{table}[H]
\caption{PER-RECORDING PERFORMANCE COMPARISON WITH
STATE-OF-THE-ART METHODS.}
\begin{adjustbox}{width=0.5\textwidth}
\begin{tabular}{ccllll}
\hline
Work         & Year & Acc(\%)      & Sens(\%)     & Spec(\%)     & Corr                     \\ \hline
Song \textit{et al.}\cite{song2015}   & 2016 & 97.1         & 95.8         & 100          & 0.860                    \\
Sharma \textit{et al.}\cite{sharma2016} & 2016 & 97.1         & 95.8         & 100          & 0.841                    \\
Li \textit{et al.}\cite{li2018}     & 2018 & 100          & 100          & 100          & \multicolumn{1}{r}{N/A} \\
Wang \textit{et al.}\cite{wang2019}   & 2019 & 97.1         & 100          & 91.7         & 0.943                    \\
Feng \textit{et al.}\cite{feng2021} & 2020 & 97.1         & 95.7         & 100          & \multicolumn{1}{r}{N/A} \\
Shen \textit{et al.}\cite{shen2021}   & 2021 & 100          & 100          & 100          & \multicolumn{1}{r}{N/A} \\
Chen \textit{et al.}\cite{chen2021}  & 2021 & 100          & 100          & 100          & 0.979                    \\
Yang \textit{et al.}\cite{yang2022}   & 2022 & 100          & 100          & 100          & 0.985                    \\
Chen \textit{et al.}\cite{chen2022}   & 2022 & 97.1         & 95.7         & 100          & 0.984                    \\
Chen \textit{et al.}\cite{chen2023}   & 2023 & 100          & 100          & 100          & 0.986                    \\
\textbf{Our[a]}          &    -  & \textbf{100} & \textbf{100} & \textbf{100} & \textbf{0.989}           \\ \hline
\end{tabular}
\end{adjustbox}
\end{table}

\subsection{Key findings and limitations}
\subsubsection{Key findings} This finding observes that incorporating more comprehensive information from the entire segment of an ECG signal-lead, by using Euclidean distance to quantify the dissimilarity of each segment with the input, significantly enhances feature extraction for CNN models. This method departs from traditional feature extraction techniques, which typically focus on R-peak and RR-interval data. The empirical results demonstrate the efficacy of the proposed approach when compared to models with similar architecture as well as other CNN methodologies. These findings represent a pivotal element of this research, shedding light on the potential of alternative feature extraction methods in improving ECG analysis and ECG classification.
\subsubsection{Limitations}

This study has several limitations that should be addressed in future work. Firstly, MPCNN was confined to the PhysioNet Apnea-ECG dataset, which comprises data from only 70 subjects. Consequently, the results may not be generalizable to other datasets, particularly those that represent a broader range of real-world scenarios with independent external data. Additionally, the proposed approach focused on distinguishing between apnea and non-apnea events, whereas hypoapnea is also a critical condition to consider in the context of SA (SA) detection \cite{Gould2012}. Collaborating with medical institutions to obtain a more varied dataset—including categories for normal, apnea, and hypoapnea—would likely enhance the robustness and applicability of the new feature extraction and CNN architecture.

Secondly, the evaluation of MPCNN was based solely on its performance with the CNN model. The Euclidean distance relationships derived during feature extraction were not extensively analyzed prior to their application in the CNN model. Further statistical analysis is warranted to deepen the understanding of these distance relationships and the potential insights they may offer. Enhancing feature extraction with methodologies based on MP algorithms could have a substantial impact on the medical field. This impact extends beyond classification to the localization of abnormal ECG segments, thereby improving the precision of medical diagnostics and treatment strategies.

\section{Conclusion}

In this study, we propose a novel feature extraction method for single-lead ECG analysis. Specifically, the proposed approach is inspired by MP algorithms, which utilize fixed-length subsequence distance profiles to capture critical features in the PQRST segment of an ECG signal. We extracted MinDP, MaxDP, and MeanDP values from these distance profiles to serve as inputs for CNN models. We compared this new feature extraction approach with conventional methods, such as R-peaks and RR intervals, in various experiments. The results demonstrate that our technique has significant potential and efficacy for SA classification, delivering promising per-segment and per-recording performance metrics. In the main experiment using the LeNet-5 model, the per-segment performance for BAFNet reached \(91.89 \%\), \(89.36 \%\), and \(93.46 \%\), in terms of accuracy, sensitivity, and specificity, respectively. Moreover, the per-segment accuracy could ascend to \(92.11 \%\), using the existing SE-MSCNN architecture. As for per-recording performance, our method achieved \(100 \%\) in all metrics and registered the highest correlation coefficient of \(0.989\). Compared to existing state-of-the-art SA detection methods, the proposed feature extraction technique outperforms them in both per-segment and per-recording metrics. This indicates that our method not only have the best best performance in state-of-art but also enhances the performance of lightweight models, thereby potentially improving the effectiveness of HSAT. Further work should involve labeling with additional properties such as hypoapnea and testing with different datasets. Additionally, analyzing the statistics of the data relationships, particularly the Euclidean distance in ECG signals, could open new avenues not only for medical classification but also for detecting abnormal events in hospitalized patients.

\begin{IEEEbiography}[{\includegraphics[width=1in,height=1.25in,clip,keepaspectratio]{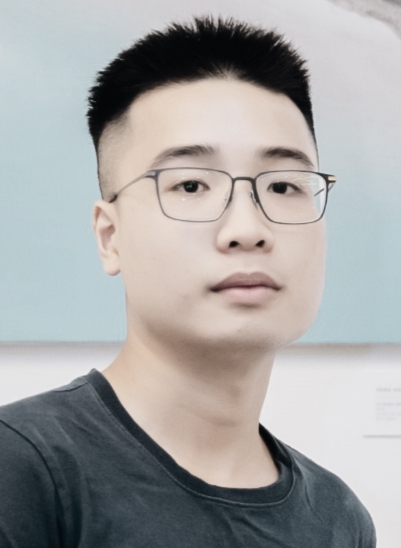}}]{Xuan Hieu Nguyen}
Hieu Nguyen graduated of the mathematics-specialized class at Nguyen Chi Thanh High School for Gifted Students in 2020. He was awarded third prize in the Vietnam Mathematical Olympiad (VMO) in the same year. Currently, he is pursuing a bachelor's degree in Electrical Engineering at VinUniversity with an interested in signal processing, embedded systems, and semiconductors. He has a keen interest research in the application of technology in the medical field and is committed to leveraging high-tech solutions to enhance the accuracy of medical diagnostics and treatment strategies.
\end{IEEEbiography}
\begin{IEEEbiography}[{\includegraphics[width=1in,height=1.25in,clip,keepaspectratio]{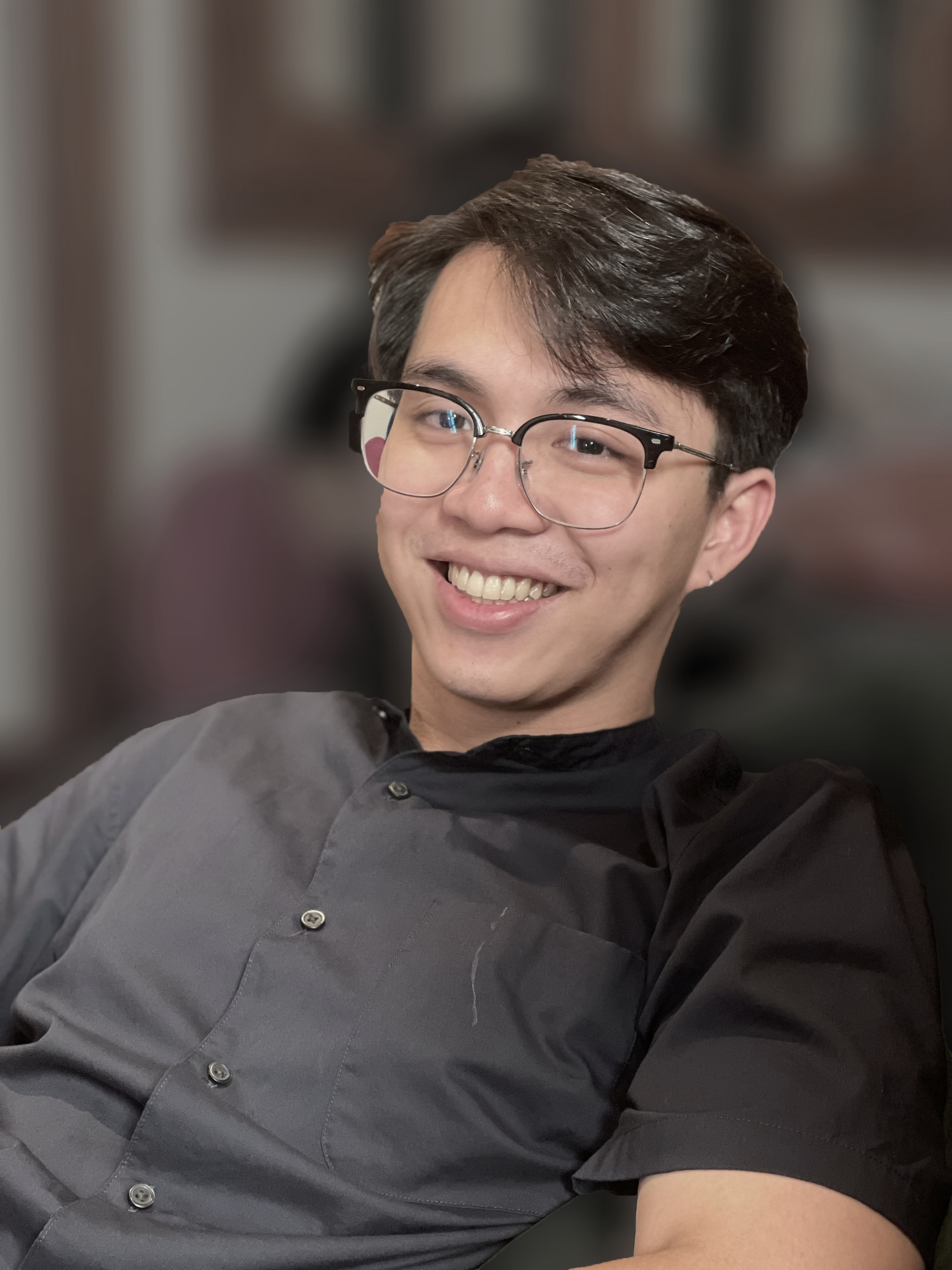}}]{Viet Duong Nguyen}
Duong Nguyen's academic journey began with a rigorous Physics curriculum at Phan Boi Chau High School for the Gifted, from which he graduated in 2021. That same year, his scholarly excellence was recognized with a third-place award in a provincial competition celebrating student achievements.At present, he is enrolled in the Electrical Engineering bachelor's program at VinUniversity, delving into areas such as embedded systems and semiconductor technology. With a keen interest in the convergence of technology and healthcare, Duong is dedicated to the advancement of integrated systems within biomedical applications. His ambition is centered on the design and optimization of integrated circuits (ICs) for biomedical devices. This endeavor aims to revolutionize the way medical diagnoses are conducted and treatments are administered, using his expertise in electrical engineering to pave the way for more efficient and accurate medical technology.

\end{IEEEbiography}
\begin{IEEEbiography}[{\includegraphics[width=1in,height=1.25in,clip,keepaspectratio]{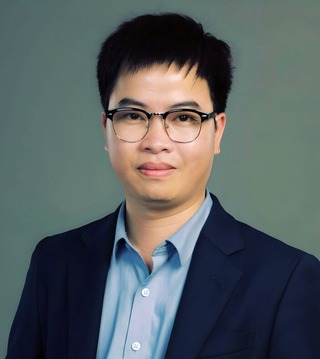}}]{Huy Hieu Pham}
Hieu Pham (Member, IEEE) received the Engineering degree in Industrial Informatics from the Hanoi University of Science and Technology (HUST), Vietnam, in 2016, and the Ph.D. degree in Computer Science from the Toulouse Computer Science Research Institute (IRIT), University of Toulouse, France, in 2019. He is currently an Assistant Professor at the College of Engineering and Computer Science
(CECS), VinUniversity, and serves as an Associate Director at the VinUni-Illinois Smart Health Center (VISHC). His research interests include computer vision, machine learning, medical image analysis, and their applications in smart healthcare. He is the author and co-author of 30 scientific articles appeared in about 20 conferences and journals, such as Computer Vision and Image Understanding, Neurocomputing, Scientific Data (Nature), International Conference on Medical Image Computing and Computer-Assisted Intervention (MICCAI), Medical Imaging with Deep Learning (MIDL), IEEE International Conference on Image Processing (ICIP), and IEEE International
Conference on Computer Vision (ICCV), Asian Conference on Computer
Vision (ACCV). He is also currently serving as Reviewer for MICCAI, ICCV, CVPR, IET Computer Vision Journal (IET-CVI), IEEE Journal of Biomedical and Health Informatics (JBHI), IEEE Journal of Selected Topics in Signal Processing, and Scientific Reports (Nature). Before joining VinUniversity, he worked at the Vingroup Big Data Institute (VinBigData), as a Research Scientist and the Head of the Fundamental Research Team. With this position, he led several research projects on medical AI, including collecting various types of medical data, managing and annotating data, and developing new AI solutions for medical analysis.
\end{IEEEbiography}

\begin{IEEEbiography}[{\includegraphics[width=1in,height=0.8in,clip]{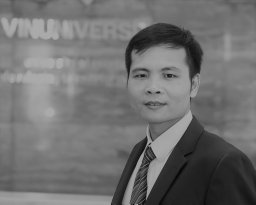}}]{Cuong D. Do}

(Member, IEEE) received his B.Sc.
degree in electronics and telecommunication from
Vietnam National University, Hanoi, Vietnam, in
2004, M.Eng. degree in electronics from Chungbuk
National University, Korea, in 2007, and a Ph.D.
degree in electronics from Cork Institute of Technol-
ogy, Ireland, in 2012. He had more than four years
working as a postdoc researcher at the University
of Cambridge, U.K in both fields of MEMS and
CMOS circuits for low-power sensors and timing
applications. He is now an Assistant Professor at
VinUniversity. His research interests include sensors for medical applications and AI for healthcare.
\end{IEEEbiography}
\end{document}